\documentclass[runningheads]{llncs}
\usepackage[T1]{fontenc}
\usepackage{graphicx,verbatim}
\usepackage{placeins}
\usepackage{hyperref}
\usepackage{xcolor}
\usepackage{dsfont}
\usepackage{amsmath}
\usepackage{booktabs} 
\usepackage{makecell}
\usepackage{tabularx}
\usepackage{cleveref}
\usepackage{multirow}
\newcolumntype{C}{>{\centering\arraybackslash}X}

\begin{document}
\title{When the Edit Changes the Patient:\\Measuring Identity Preservation in Counterfactual Retinal Images}
\titlerunning{Measuring Identity Preservation in Counterfactual Retinal Images}
\author{
    Andrea Posada\inst{1,2}$^\star$\orcidID{0009-0005-3021-5525} \and
    Wenke Karbole\inst{1,3}$^\star$ \and
    Bach Ngoc Doan\inst{1}$^\star$ \and
    Alexander Weers\inst{1,2} \and
    Solmaz Abdolrahimzadeh\inst{4} \and
    Maria Patsiamanidi\inst{5} \and
    Kahkashan Haider\inst{6} \and
    Vaishali Khare\inst{6} \and
    Daniel Rueckert\inst{1,2,7} \and
    Andrew Lotery\inst{5} \and
    Sobha Sivaprasad\inst{6} \and
    Martin J. Menten\inst{1,2,7}
}
\authorrunning{Posada, Karbole, Doan et al.}
\institute{
    Technical University of Munich and TUM University Hospital, Germany  \\
    \email{wenke.karbole@tum.de} \and
    Munich Center of Machine Learning (MCML), Munich, Germany \and
    Munich Data Science Institute (MDSI), Munich, Germany \and
    Sapienza University of Rome, Rome, Italy \and
    Faculty of Medicine, University of Southampton, Southampton, United Kingdom \and
    Moorfields Eye Hospital NHS Foundation Trust, London, United Kingdom \and
    Imperial College London, UK \\
    $^\star$ Equal contribution.
}

\maketitle

\begin{abstract}
Counterfactual medical image generation aims to modify an existing image to reflect a hypothetical scenario in which certain characteristics of the imaged subject are altered, while keeping their identity fixed. Most existing works repurpose established image editing methods, which do not directly supervise identity preservation. Instead, they assume that identity is implicitly preserved by anchoring generation to the source image. This assumption is rarely tested and may fail in domains where biometric cues are subtle, such as retinal optical coherence tomography (OCT). In this work, we explicitly measure identity preservation for three groups of text-conditioned editing methods -- source-anchored, structured-prompt, and paired-training -- using referee classifiers, embedding alignment scores, and a blind reader study. We find that all methods produce high-quality OCT images with comparable editing success, yet their identity preservation differs markedly. Source-anchored editing frequently alters the depicted subject, while paired-training preserves it best. We argue that future work on medical counterfactual generation must explicitly measure and report identity preservation alongside image realism and editing success.
\keywords{Counterfactual image generation \and Text-conditioned image editing \and Identity preservation \and Retina \and OCT imaging}

\end{abstract}

\section{Introduction}
\label{sec:intro}

Counterfactual medical image generation aims to answer the question: \emph{``What would this subject's image look like if a given attribute were changed?''} In order to be useful for applications such as visualizing disease progression and simulating treatment effects~\cite{yeganeh2025latent,gu2023biomedjourney,prism2025}, the resulting synthetic images must not only be realistic and anatomically accurate, but also preserve the subject's identity.

\begin{figure}[t]
\centering
\includegraphics[width=\textwidth]{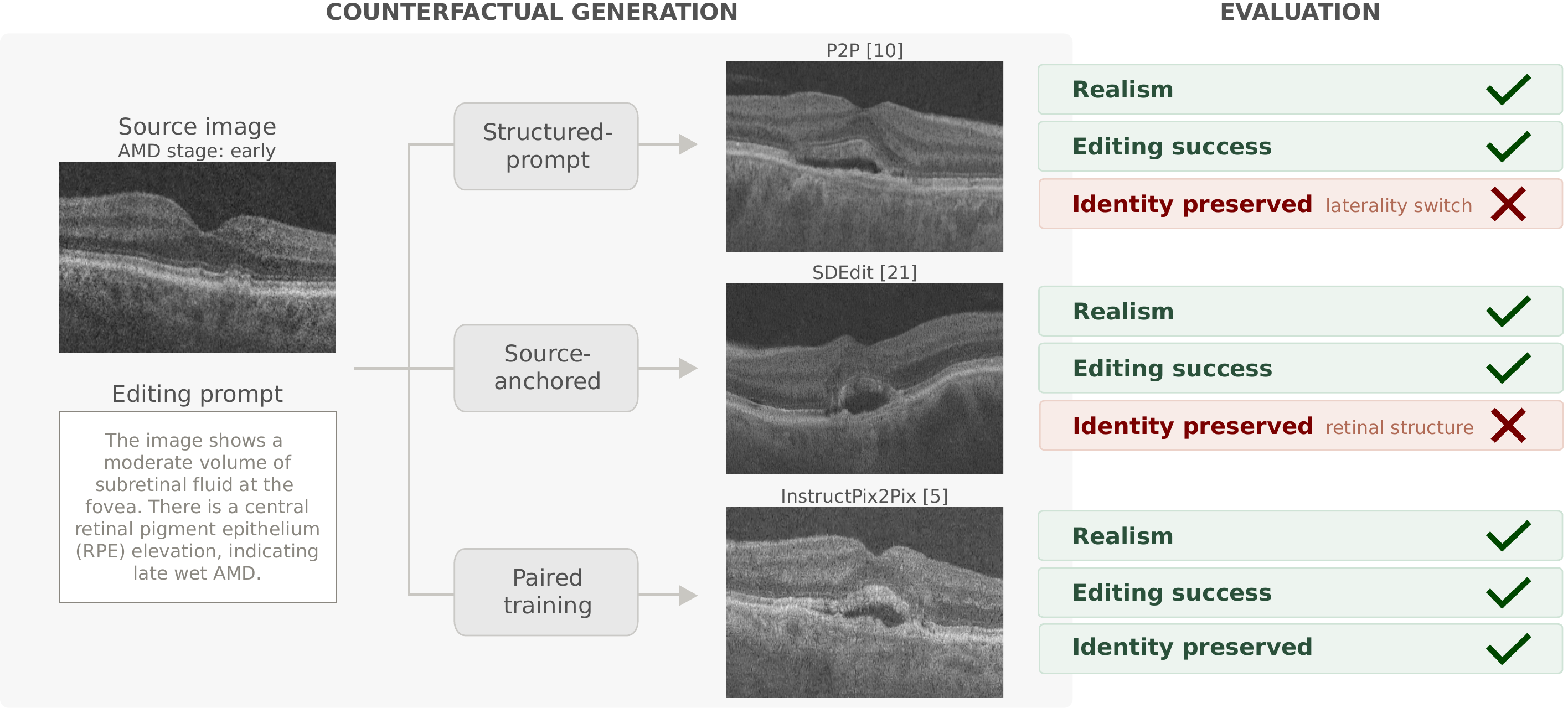}
\caption{In OCT counterfactual generation, changing global age-related macular degeneration (AMD) biomarkers requires strong edits that can disrupt the subtle cues linked to patient identity. As a result, the generated samples often fail to preserve the subject's identity even when the edits appear faithful and realistic.
}
\label{fig:identity_preservation}
\end{figure}

Most existing text-conditioned counterfactual methods repurpose established diffusion-based image editing techniques, which do not directly supervise identity preservation. Instead, they assume it is preserved implicitly, by anchoring generation to the source image or by training on image pairs. This assumption has largely held because most medical counterfactual work has been developed and evaluated on chest X-rays~\cite{gu2023biomedjourney,perezgarcia2024radedit,chambon2022roentgen}, where the subject's identity is strongly linked to salient biometric features: skeletal structure, body habitus, heart outlines and lung shape vary markedly between subjects and enable near-perfect patient re-identification~\cite{packhauser2022reid}. Conversely, in retinal optical coherence tomography (OCT), identity cues are encoded by more subtle features such as retinal-layer thickness and blood vessels~\cite{kong2020heritability}. We hypothesize that, for OCT images, even small edits can substantially alter a subject's biometric features and thus produce implausible counterfactuals. Consequently, the assumption of identity preservation as an automatic by-product of current text-conditioned editing methods may not hold in retinal imaging (\Cref{fig:identity_preservation}).

To test this hypothesis, we conduct experiments using a longitudinal OCT dataset of subjects with age-related macular degeneration (AMD). Global AMD-related changes often superimpose subtle identity cues, making counterfactual editing challenging. We compare three text-conditioned image editing groups -- source-anchored, structured-prompt, and paired-training -- that have not been optimized for identity preservation. In addition to realism and editing success, we assess identity preservation using three complementary approaches: referee classifiers for the subject sex and eye identity, an alignment score between the generated edit and the subject's real longitudinal change, and a blind reader study in which clinically trained readers judge whether image pairs depict the same eye.

Our findings show that identity preservation cannot be taken for granted: even small edits that appear realistic and achieve the intended change can alter the depicted subject's identity. Moreover, accounting for it during model development informs hyperparameter selection at little cost to realism or editing success. We therefore argue that identity preservation in medical counterfactual generation must be explicitly measured rather than assumed, particularly for imaging modalities with subtle identity cues such as retinal OCT.

\section{Related Work}
\label{sec:relwork}

\subsection{Text-Conditioned Medical Counterfactual Generation}
\label{sec:relwork_methods}
Text-conditioned counterfactual generation enables precise manipulation of specific pathological or demographic traits through language. Most existing work repurposes established text-conditioned diffusion-based image editing methods \cite{ilanchezian2023retinal,gu2023biomedjourney,prism2025,yoon2023sadm,perezgarcia2024radedit}. These methods can be grouped into four categories according to how they modify the underlying source image: source-anchored, structured-prompt, text- and mask-conditioned, and paired-training. Across all four groups, identity preservation is usually treated as a by-product of the editing mechanism rather than an explicit objective.

Source-anchored methods closely tie the diffusion process to the source image, so that identity is retained through proximity to the source alone. SDEdit~\cite{meng2021sdedit}, one of the most representative works, partially noises the source and denoises it under the new prompt. Although training-free and backbone-agnostic, these methods offer limited editing control. Classifier-guided variants~\cite{jeanneret2022dime,augustin2022dvce,jeanneret2023ace,ilanchezian2023retinal}, which steer denoising with classifier gradients, provide greater control but are restricted to a classifier's label space rather than free-form text.

Structured-prompt methods predominantly build on Prompt-to-Prompt~\cite{hertz2022prompt} via null-text inversion~\cite{mokady2023nulltext}: an image is edited by manipulating the word-level attention maps of the source and target text prompts during the generation process. Since such edits typically do not pertain to identity, it is expected to be preserved implicitly. Although this paradigm has been applied to medical counterfactual generation~\cite{prism2025}, it relies on closely aligned prompts that differ by only a few words, whereas clinical data consists of free-text reports and multi-word pathology descriptions.

Text- and mask-conditioned methods~\cite{avrahami2022blended,couairon2023diffedit,perezgarcia2024radedit,fontanella2024diffusion} confine changes to a user-provided or automatically estimated region, preserving structures outside the mask by design. Nonetheless, they require accurate masks and cannot model edits with global effects such as aging, sex, or diffuse pathology. Because most AMD-related changes are inherently global, impacting multiple retinal layers, we do not investigate these strategies further in our study.

Paired-training methods learn counterfactual edits from same-subject image pairs. InstructPix2Pix~\cite{brooks2023instructpix2pix} popularized this concept for natural images, while BiomedJourney~\cite{gu2023biomedjourney} and SADM~\cite{yoon2023sadm} extend it to longitudinal medical data. Paired data, however, is not readily available in medical imaging.

\subsection{Measuring Identity Preservation}
\label{sec:relwork_identity}
Although identity preservation is central to a plausible counterfactual, it is rarely measured directly. Most work instead reports \emph{proximity} metrics~\cite{prism2025,melistas2024benchmarking,yeganeh2025latent,monteiro2023axiomatic}, such as pixel-level L1 distances, structural similarity~\cite{ssim}, or perceptual distance~\cite{lpips}. These quantify how much an image changes as a result of an edit, thereby conflating identity with \emph{minimality}. As a consequence, they are insensitive to small changes that specifically alter subject identity, while large but identity-consistent edits are overly penalized. Only a few works report attribute preservation in their evaluation~\cite{menten2023exploring,xia2024mitigating,min2025instructx2x}, and methods that account for identity preservation directly during training are rarely adopted in practice~\cite{maeng2025idenbat,xia2021learning}. Identity preservation thus remains an overlooked aspect of recent work on medical counterfactual image generation.

\section{Method}
\label{sec:method}
We investigate to which extent image editing preserves subject identity when generating counterfactual images on a longitudinal OCT dataset of AMD patients (\Cref{sec:dataset}). To this end, we employ three complementary evaluation approaches -- referee networks, embedding-alignment metrics, and a blind reader study (\Cref{sec:metrics}) -- to compare three image editing strategies: source-anchored, structured-prompt, and paired-training (\Cref{sec:models}).

\subsection{Dataset}
\label{sec:dataset}
All experiments were performed on a in-house dataset from the University Hospital Southampton containing 43\,165 OCT images from 6\,157 eyes of AMD patients. Most patients contributed multiple scans over a time period of up to seven years, that may span their eye's conversion from early and intermediate AMD to late stages of the disease. For each image, we used the central B-Scan and automatically generated a text description using a vision language model for ophthalmological report generation~\cite{retinavlm_full}.The dataset is split at the subject level into training (80\%), validation (10\%), and test (10\%) sets. To guarantee consistency between the report and expert-derived ground truth AMD stage labels, we extracted the stage from the report using both a regex- and LLM-based approach, and disagreements were resolved by a human annotator. Reports that did not mention an AMD stage or whose staging was inconclusive were excluded from the validation and test sets.

\subsection{Evaluation Metrics}
\label{sec:metrics}
We quantitatively measured multiple properties of a generated counterfactual $x_g$. $x_g$ is generated from a real image $x_s$ conditioned on a text prompt $p_t$, which describes the intended counterfactual scenario. We may also consider a real target image $x_t$ as one valid counterfactual out of many. Notice that $p_t$ is the textual report describing $x_t$.

\noindent\textbf{Identity Preservation via Referee-Based Methods}
Since edits are restricted to AMD-related findings, all other attributes should remain unchanged. We thus assess identity preservation with two referee classifiers and a blind reader study, rating whether two images (real/edited or real/real) depict the same eye.

As the primary referee, a Siamese-based \emph{same-eye verifier} $f_\text{eye}$ was trained to predict whether two OCT scans originate from the same subject and laterality (F1-score of 0.97). Following biometric verification, the identity agreement between $x_s$ and $x_g$ is defined as:
\begin{equation}
  \text{Agr}_{f_\text{eye}}
    = \frac{1}{|A|}\sum_{i \in A}
      \mathds{1}{\left[f_\text{eye}\left(x_g^{(i)}, x_s^{(i)}\right) \geq \tau_{0.1}\right]}
\end{equation}
with $A = \left\{i : f_\text{eye}\left(x_s^{(i)}, x_t^{(i)}\right) \geq \tau_{0.1}\right\}$ to correct for the classifier's unreliability. The threshold $\tau_{0.1}$ is selected on the validation split such that the false accept rate on different-eye pairs does not exceed 0.1\%.

As an additional referee, we considered a binary \emph{sex classifier} $f_\text{sex}$\textcolor{blue}, which serves as a weaker but complementary identity proxy. It predicts the sex $\hat{s}$ of a subject from the OCT B-scan, an attribute ideally unaffected by editing (F1-score of 0.71). We measure the agreement between the sex predictions on $x_g$ and $x_s$ via:
\begin{equation}
  \text{Agr}_{f_\text{sex}}
    = \frac{1}{|B|}\sum_{i \in B}
      \mathds{1}{\left[\hat{s}_g^{(i)} = \hat{s}_s^{(i)}\right]}
\end{equation}
where $B = \left\{i : \hat{s}_s^{(i)} = \hat{s}_t^{(i)}\right\}$ accounts for the classifier's unreliability. Both referee classifiers were trained, validated, and tested on the same data splits used for the generative editing models to avoid data leakage.

We also conducted a \textit{reader study} in which four clinically trained readers judged whether a pair of OCT images showed the same eye. Each reader assessed 240 randomly sampled pairs: 60 pairs per editing method plus 60 pairs of real images as a control. Within each subset, half of the pairs showed the same eye and half showed different eyes. Subjects were selected at random, and readers were blinded to the editing method.

\noindent \textbf{Editing Success and Identity Preservation via Embedding Alignments} 
To measure the alignment between $x_g$ and $p_t$, we report their mean-corrected CLIP-score~\cite{hessel2021clipscore,biomedclip}. We further define the \emph{image change alignment} (ICA) to measure whether an edit moves $x_s$ in the same semantic direction as the real longitudinal transition from $x_s$ to $x_t$ in the latent space. Let $e_s$, $e_t$, and $e_g$ denote image embeddings using Zhang et al.'s model~\cite{biomedclip}. Then:
\begin{equation}
  \text{ICA}
  = \frac{1}{N}\sum_{i=1}^{N}
    S_c\left(e_t^{(i)} - e_s^{(i)},\ e_g^{(i)} - e_s^{(i)}\right)
\end{equation}
where $S_c$ is the cosine similarity and $N$ denotes the number of samples. A score near~1 indicates that the edit is semantically aligned with the real subject-specific transition. Because real transitions preserve identity by definition, close alignment indicates that the editing operation is not only successful but also maintains the subject's identity.

\noindent \textbf{Realism and Editing Success Metrics}
We measure the realism of the generated edits using the Fr\'{e}chet Inception Distance (FID)~\cite{fid}. Editing success is also evaluated via the F1-score and the attribution score~\cite{nemirovsky2021countergan} using a multi-class AMD stage classifier $f$ (F1-score of 0.57). We use the attribution score, also known as prediction gain, to measure how the predicted logits of the target AMD class $c_t$ change with the edit. While standard formulations focus only on cases where the class changes between source and target, we extend the metric to also handle same-stage edits:
\begin{equation}
  \phi = \frac{1}{N}\sum_{i=1}^{N} \operatorname{sign}\left(
    f\left(x_t^{(i)}\right)_{c_t^{(i)}} - f\left(x_s^{(i)}\right)_{c_t^{(i)}}
  \right)
  \cdot
  \left(
    f\left(x_g^{(i)}\right)_{c_t^{(i)}} - f\left(x_s^{(i)}\right)_{c_t^{(i)}}
  \right).
  \label{eq:predgain}
\end{equation}

\subsection{Counterfactual Image Generation Strategies}
\label{sec:models}
We base all investigated strategies on MediSyn~\cite{medisyn}, a generalist medical text-to-image latent diffusion model, and fine-tune it on text-image pairs of the OCT dataset. Since medical terminology differs markedly from natural language, we trained MediSyn with BioMedCLIP~\cite{biomedclip} and BioLORD~\cite{biolord} text encoders and selected the best-performing variant per method.

\noindent \textbf{Source-anchored -- Stochastic Differential Editing (SDEdit)}
SDEdit~\cite{meng2021sdedit} is applied at inference time. Rather than initializing the denoising process from pure noise, it starts from a noisy version of $x_s$. The strength hyperparameter $\gamma$ sets the fraction of steps for the noising and denoising processes. We tuned $\gamma$ over \{0.2, 0.4, 0.6, 0.8, 1.0\}.

\noindent \textbf{Structured-prompt -- Prompt-to-Prompt (P2P)}
P2P~\cite{hertz2022prompt} is also applied at inference time. First, null-text inversion~\cite{mokady2023nulltext} inverts $x_s$ together with its original prompt, recovering a denoising trajectory that reproduces $x_s$ from noise. P2P then denoises along this trajectory conditioned on $p_t$, but it injects the attention maps of the source prompt instead of the target's for a fraction of the steps. This fraction is determined by the self-replacement rate $srs$. Reusing these attention maps anchors the generation process to $x_s$. We tuned $srs$ over \{0.8, 0.6, 0.4, 0.2, 0.0\}. Although P2P is designed to modify single words in otherwise identical prompts and our free-text reports differ throughout, we find that it produces meaningful counterfactual images, competitive with the other methods.

\noindent \textbf{Paired training -- InstructPix2Pix}
Unlike the inference-time strategies SDEdit and P2P, \emph{InstructPix2Pix}~\cite{brooks2023instructpix2pix} further trains the OCT-fine-tuned MediSyn model to learn semantic editing explicitly from paired data. The model is conditioned jointly on $p_t$ and $x_s$ to predict $x_t$. During inference, the edited image is obtained via a single conditioned denoising pass with separate classifier-free guidance scales for the image ($igs$) and the text ($gs$). We tuned $igs$ as the main parameter for feature preservation over \{1.0, 1.5, 2.0, 2.5, 3.0\} and $gs$ over \{5.5, 7.5, 11.5\}. Whereas the original InstructPix2Pix is trained on synthetically generated pairs, we use real source-target pairs of the same subject.

Two configurations per editing strategy are chosen by the best average rank across the relevant metrics: an \emph{edit-focused} selection, ranking on realism and editing success only, and a \emph{balanced} selection, which additionally accounts for identity preservation (\Cref{sec:experiment1}). We then analyze how each metric varies with the methods' respective editing strength hyperparameters (\Cref{sec:experiment2}).

\section{Results and Discussion}
\label{sec:results}

\subsection{Measuring Identity Preservation in Image Editing}
\label{sec:experiment1}

\begin{figure}[t]
    \centering
    \includegraphics[width=0.95\linewidth]{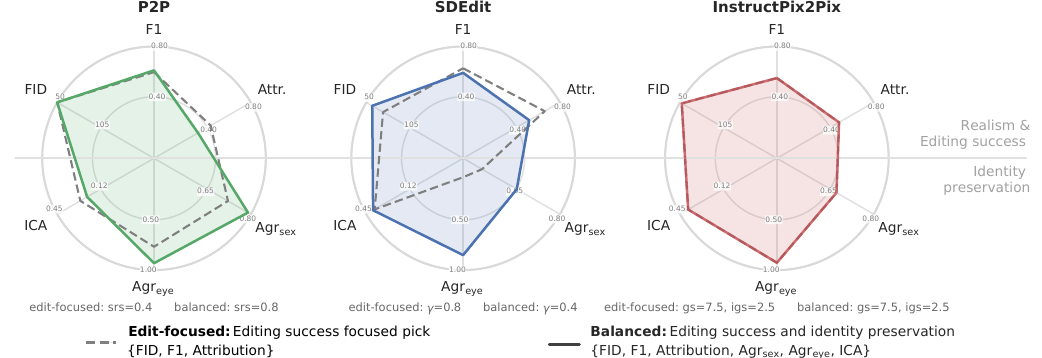}
    \caption{Per-method comparison on two selection criteria across realism, editing success, and identity preservation metrics. Selecting on realism and editing success alone (dashed line) can yield configurations with substantial loss of identity features. The balanced pick (solid line) recovers identity preservation while editing success and realism are largely retained.}
    \label{fig:modelPicks}
\end{figure}

\begin{figure}[t]
    \centering
    \includegraphics[width=\linewidth]{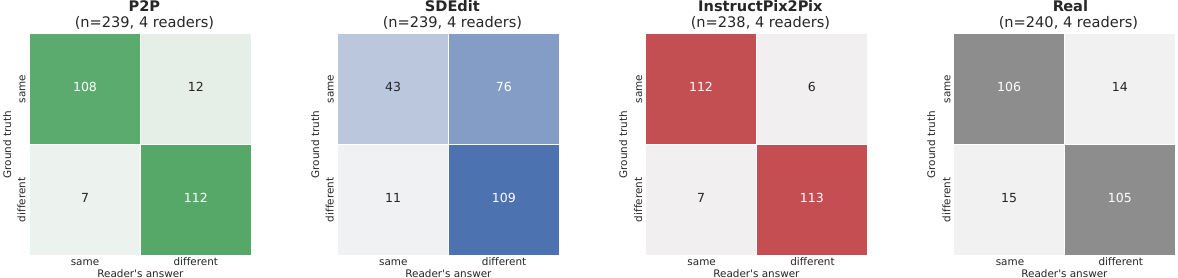}
    \caption{Two displayed images (real/real) or (real/edited) were graded by four clinically trained readers as the same or different eyes with substantial agreement (Fleiss's $\kappa=0.67$). While edits generated by P2P and InstructPix2Pix are mostly recognized as belonging to the same eye, those of SDEdit reveal frequent identity loss.}
    \label{fig:readerStudy}
\end{figure}

\begin{figure}[t]
    \centering
    \includegraphics[width=\linewidth]{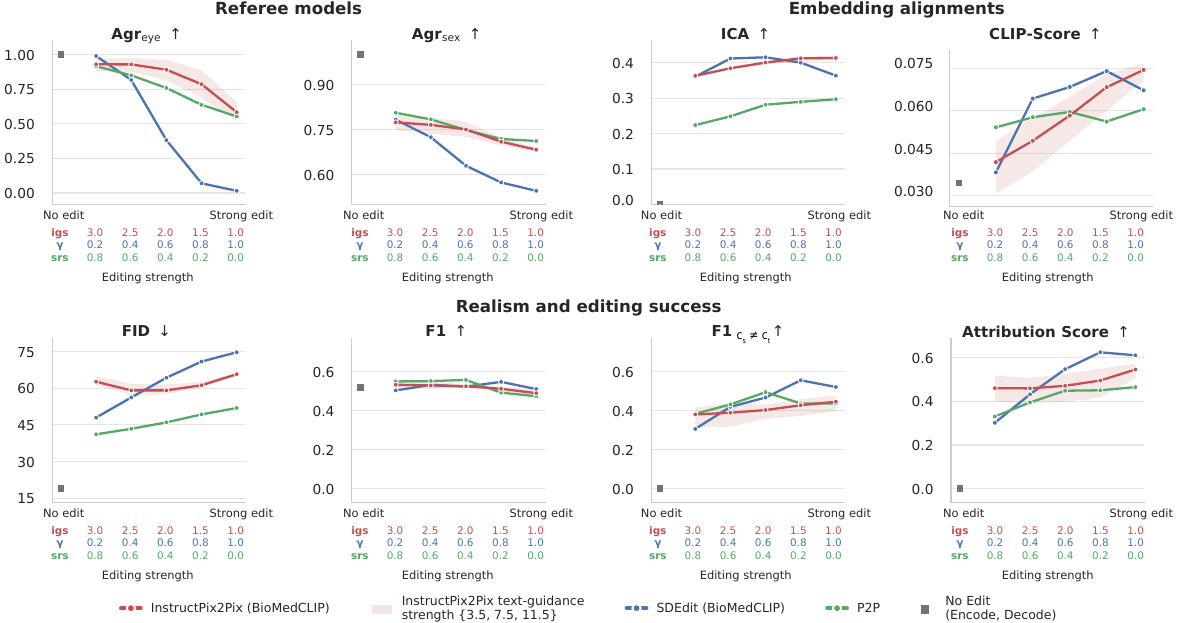}
    \caption{With increasing editing strength, all investigated text-conditioned editing strategies show a loss in identity preservation, most prominently revealed by $\text{Agr}_\text{eye}, \text{Agr}_\text{sex}$. Above all, we find that even minor edits can cause a drastic loss of identity features, which is not equally compensated by gains in editing success metrics.}
    \label{fig:dynamics}
\end{figure}

For two out of three image editing methods, the configurations selected by the edit-focused and balanced criteria meaningfully differ (\Cref{fig:modelPicks}). For P2P and SDEdit, the balanced criterion favors a lower editing strength than the edit-focused criterion. This changes realism and editing success only marginally, while improving eye-identity and sex preservation, as measured by the referee classifiers. The most severe identity loss is observed for the source-anchored method SDEdit under the edit-focused criterion. Qualitative results confirm a high overall image quality and realism (\Cref{fig:identity_preservation}). The reader study supports our findings: edits from the edit-focused SDEdit model were more often perceived as different eyes than as the same (\Cref{fig:readerStudy}). For P2P and InstructPix2Pix, however, reader performance was comparable to the performance on real image pairs. Overall, these results show that selecting models based on realism and editing success alone does not guarantee that subject identity is preserved.

\subsection{Editing Strength as Decisive Hyperparameter}
\label{sec:experiment2}

All methods show a clear negative correlation between identity preservation metrics and the editing strength hyperparameter (\Cref{fig:dynamics}). The F1-score remains roughly constant while the attribution score rises, which is explained by our test-set composition: most transitions in our data are subclinical, with only 479 cases (26\%) involving an actual AMD-stage change. Restricted to these cases, the F1-score accordingly exhibits the expected upward trend. The FID trend is consistent with prior work~\cite{augustin2022dvce,DVCE_7}, in that lower editing strength keeps generated images closer to the real-data distribution.  Since all editing strength hyperparameters are method-specific, the results are not directly comparable across methods. Within their respective ranges, however, InstructPix2Pix and P2P show a high robustness to variations in editing strength. This is consistent with \Cref{sec:experiment1}, where they exhibited the strongest implicit identity signal.

\section{Conclusion}

This study has evaluated whether text-conditioned image editing methods, used for counterfactual generation, preserve subject identity when applied to retinal OCT images. While all three investigated methods produce high-quality OCT images with comparable editing success, some methods alter subtle biometric features of the eye. Therefore, identity preservation cannot be assumed as an automatic by-product of anchoring the generative process to the source image. 

In response, we have demonstrated that evaluating generative models solely through the lens of image realism and editing success does not reliably capture changes in identity, yielding invalid conterfactuals. By explicitly accounting for identity preservation already during model selection and tuning, one can substantially improve this property with minimal deductions to image realism and editing success. Consequently, we argue that future work on medical counterfactual generation must explicitly measure and report identity preservation alongside image realism and editing success, particularly for imaging modalities with subtle identity cues such as retinal OCT.

\begin{credits}
\subsubsection{\ackname} This work was partially supported by the German Research Foundation (Project 532139938) and EPSRC grant EP/Y015665/1.

\subsubsection{\discintname}
The authors have no competing interests to declare that are
relevant to the content of this article.
\end{credits}

%
%
\bibliographystyle{splncs04}
\bibliography{Paper-0010}
\end{document}